\documentclass[cameraready]{Interspeech}

\usepackage{algpseudocode}
\usepackage{algorithm}
\usepackage{multirow}
\usepackage{csquotes}
\usepackage[T1]{fontenc}
\usepackage{subcaption}
\usepackage[dvipsnames]{xcolor}

\title{Pretrained self-supervised speech models can recognize unseen consonants}

\author[affiliation={1}, orcid=0009-0009-3690-9412, correspondingauthor, equalcontribution]{Chihiro}{Taguchi}
\author[affiliation={2}, orcid=0009-0001-0363-6771, equalcontribution]{Éric}{Le Ferrand}
\author[affiliation={3}, orcid=0009-0006-4938-7391]{Hirosi}{Nakagawa}
\author[affiliation={4}, orcid=0000-0002-8899-6679]{Hitomi}{Ono}
\author[affiliation={3}, orcid=0000-0001-8177-6182]{Kanji}{Kato}
\author[affiliation={5}, orcid=0000-0003-3318-892X]{Emily}{Prud'hommeaux}
\author[affiliation={1}, orcid=0000-0002-0435-4864]{David}{Chiang}

\newcommand{\authorseplen}[0]{0.7em}
\address{
    $^1$ University of Notre Dame, USA \hspace{\authorseplen}
    $^2$ University at Buffalo, USA \\
    $^3$ Tokyo University of Foreign Studies, Japan \hspace{\authorseplen}
    $^4$ Reitaku University, Japan \hspace{\authorseplen}
    $^5$ Boston College, USA
}

\email{
    ctaguchi@nd.edu,
    ericlefe@buffalo.edu,
    nhirosi@tufs.ac.jp,
    ono@reitaku-u.ac.jp,
    jiateng.ganzhi@gmail.com,
    prudhome@bc.edu,
    dchiang@nd.edu
}

\usepackage{comment}

\newcommand{\dc}[0]{{\textpipe}} %
\newcommand{\ac}[0]{\textipa{!}} %
\newcommand{\pc}[0]{{\textdoublebarpipe}} %
\newcommand{\lc}[0]{{\textdoublepipe}} %
\newcommand{\ipag}[0]{\textscriptg} %
\newcommand{\h}[0]{\textipa{\super h}}
\newcommand{\Ng}[0]{\textipa{N}}
\newcommand{\X}[0]{\textchi}
\newcommand{\qg}[0]{\textscg}
\newcommand{\orth}[1]{\guilsinglleft#1\guilsinglright}

\begin{document}

\maketitle

\begin{abstract}
    Modern pretrained self-supervised automatic speech recognition models are trained on large-scale audio data to encode speech into contextualized representations.
    However, their training data are heavily skewed toward high-resource languages with little data from low-resource languages, raising concerns about the potential underrepresentation of typologically uncommon speech sounds such as click consonants primarily found in Khoisan languages.
    This leads to our central research question: Can these models recognize click consonants as accurately as other speech sounds?
    To address this question, we fine-tune and compare pretrained self-supervised speech models (Wav2Vec2 and HuBERT) on data from two click-rich Khoisan languages (G{\textpipe}ui and West \textipa{!}Xoon).
    Our results reveal that the fine-tuned models consistently recognize clicks more accurately than non-clicks, suggesting that self-supervision enables generalization across human speech sounds including rare phonemes.
\end{abstract}

\section{Introduction}
In recent years, automatic speech recognition (ASR) has made remarkable progress in expanding multilingual capabilities and improving adaptability to under-resourced languages.
Large-scale pretrained self-supervised speech models have substantially improved performance across many languages by learning general acoustic representations from massive amounts of unlabeled multilingual audio.
Models based on self-supervised objectives have become the foundation of many state-of-the-art ASR systems, demonstrating strong transfer capabilities, especially when fine-tuned on limited labeled data.

Despite these advances, the benefits of modern ASR systems remain unevenly distributed.
The multilingual corpora used to pretrain contemporary self-supervised models are typically dominated by high-resource languages, particularly English and other widely spoken languages.
In contrast, most under-resourced languages are minimally represented or entirely absent from pretraining data.
This imbalance raises concerns about whether pretrained models adequately capture the full diversity of human speech, especially typologically uncommon phonetic phenomena.

One particularly striking case involves click consonants, which are found primarily in Khoisan languages and in some neighboring Bantu languages in Southern Africa.
Click consonants exhibit articulatory and acoustic properties that are rare in the world's languages and are virtually absent from high-resource languages that dominate ASR pretraining datasets.
Phoible \cite{phoible} reports only 12 languages to have the most crosslinguistically attested click consonant [k{\textpipe}] (tenuis dental click), of which only Zulu, Xhosa, and Northern Ndebele have been integrated in pretraining of the major ASR models (cf. Table~\ref{tab:models}).
As a result, it remains unclear whether self-supervised multilingual speech models trained predominantly on non-click languages can robustly represent and recognize these sounds.

Languages with click consonants have been neglected in technology so much that the clicks are sometimes suppressed as ``noises'' in video-conferencing tools with noise reduction (Hirosi Nakagawa, personal communication, 2025).
If speech technologies are to truly support speaking together across linguistic communities, they must function not only for globally dominant languages but also for languages with distinctive phonological systems.
Understanding how modern ASR systems handle typologically rare sounds is therefore both a technical and a social imperative.

In this work, we investigate whether pretrained self-supervised multilingual speech models can accurately recognize click consonants.
We fine-tune several widely used self-supervised architectures on data from G{\textpipe}ui and West \textipa{!}Xoon (West Taa), Khoisan languages with exceptionally rich inventories of click consonants.
We then compare recognition performance between click and non-click phonemes to assess whether typologically uncommon sounds are disadvantaged in these models.

The contributions of this study are as follows:\footnote{Part of the datasets, the trained models, and the code used in the experiments will be publicly available.}
\begin{itemize}
    \item \textbf{Dataset construction for a click-rich Khoisan languages.}
    We construct ASR datasets for the G{\textpipe}ui and West \textipa{!}Xoon languages.
    \item \textbf{Systematic evaluation of click recognition in self-supervised ASR.}
    We fine-tune multiple pretrained multilingual self-supervised ASR models and compare recognition performance between click and non-click phonemes as well as their overall performance.
    \item \textbf{Empirical evidence of robust generalization to typologically rare sounds.}
    We find that fine-tuned models consistently recognize click consonants more accurately than non-click phonemes, suggesting that self-supervised pretraining supports generalization beyond dominant language distributions.
\end{itemize}

\section{Related Work}
The introduction of Transformer \cite{vaswani2023attentionneed} has enabled end-to-end training of ASR models, with scalability to a large amount of training data and to multilingual tasks.
Roughly speaking, there are currently two architectural approaches to multilingual ASR:
(1) encoder-only self-supervised training followed by language-specific fine-tuning and
(2) encoder-decoder supervised training whose objective is to \emph{translate} speech into a sequence of tokens.

The representative of the former type of architecture is Wav2Vec 2.0 \cite{baevski2020wav2vec20frameworkselfsupervised}.
The pretraining is similar to that of BERT \cite{devlin-etal-2019-bert}.
The data consists only of unlabelled audio, and some frames are masked and quantized to discrete units that represent an abstract speech category in the learnable matrix (codebook).
For each frame, the chosen category is represented as a vector (codevector), and the objective of the training is to predict contextualized vectors through Transformer blocks such that it minimizes contrastive loss between the predicted vectors and codevectors while encoraging the model to use a diverse set of codevectors.
The pretrained model can be fine-tuned to solve an ASR task by stacking a Connectionist Temporal Classification (CTC) \cite{10.1145/1143844.1143891} layer to predict a symbol per frame and obtain the final string by collapsing the consecutive duplicate symbols.
Since its debut, several offshoot models of Wav2Vec 2.0 have been released with a growing number of languages and increasing quantities of speech data \cite{conneau2020unsupervisedcrosslingualrepresentationlearning, babu2021xlsrselfsupervisedcrosslingualspeech, communication2023seamlessmultilingualexpressivestreaming, pratap2023scalingspeechtechnology1000, omnilingualasrteam2025omnilingualasropensourcemultilingual}.
Another similar architecture of this type is HuBERT \cite{hsu2021hubertselfsupervisedspeechrepresentation}, in which the self-supervised objective is similar but creates the discrete pseudolabels (\textit{i.e.} abstract categories) of the masked frames by means of $k$-means clustering on the Mel-Frequency Cepstral Coefficients of the input audio.
A successful instance of the latter type is Whisper \cite{radford2022robustspeechrecognitionlargescale} whose training schema is similar to that of machine translation which aims to minimize the token-level cross-entropy loss.
The audio input is represented as vectors through the encoder, which is then fed into the decoder through cross-attention to predict text tokens auto-regressively.

The self-supervised speech models have been reported to have learned universal speech representation \cite{millet2022selfsupervisedspeechmodelsdevelop} and to encode language-agnostic phonetic information \cite{choi24b_interspeech}, thereby being robust in adapting to unseen languages \cite{rouditchenko2023comparisonmultilingualselfsupervisedweaklysupervised}.
However, while it is generally believed that adding more languages in pretraining data helps cross-lingual transfer \cite{grosman2025crosslingualtransferabilitypretrainedwav2vec2based}, whether these models also show robustness to individual unseen speech sounds has remained unknown.
We aim to provide an answer to this question through experiments on languages with typologically atypical consonants---click consonants.

\section{Data}
This section provides a description of the constructed datasets for G{\textpipe}ui and West \textipa{!}Xoon as well as their linguistic background.
These two languages both fall into the category of Khoisan languages famous for click consonants but belong to genealogically separate language families, Khoe--Kwadi and Tuu, respectively.
Table~\ref{tab:dataset} provides a description of the G{\textpipe}ui and West \textipa{!}Xoon datasets.

\begin{table}[t]
    \centering
    \setlength{\tabcolsep}{3.1pt}
    \caption{Dataset description.
    A \emph{word} is a unit segmented by whitespace.}
    \label{tab:dataset}
    \begin{subtable}[t]{\linewidth}
        \caption{G{\textpipe}ui.}
        \label{tab:data-gui}
        \centering
        \begin{tabular}{@{}lrrr@{}} \toprule
             & Train & Test & Total \\ \midrule
            \#Samples & 3691 & 411 & 4102 \\ \midrule
            Total length (s) & 18616 & 2044 & 20660 \\
            Avg. length (s) & 5.04 ($\pm$2.41) & 4.97 ($\pm$2.25) & 5.04 ($\pm$2.39) \\ \midrule
            Total \#words & 49058 & 5499 & 49068 \\
            Avg. \#words & 13.29 ($\pm$6.85) & 13.38 ($\pm$6.72) & 13.30 ($\pm$6.84) \\
             \bottomrule
        \end{tabular}
        \vspace{1em} %
    \end{subtable}
    \begin{subtable}[t]{\linewidth}
        \caption{West \textipa{!}Xoon.}
        \label{tab:data-west-xoon}
        \centering
    \begin{tabular}{@{}lrrr@{}} \toprule
         & Train & Test & Total \\ \midrule
        \#Samples & 864 & 246 & 1110 \\ \midrule
        Total length (s) & 5006 & 1410 & 6416 \\
        Avg. length (s) & 5.79 ($\pm$2.99) & 5.73 ($\pm$2.70) & 5.78 ($\pm$2.93) \\ \midrule
        Total \#words & 9073 & 2507 & 11580 \\
        Avg. \#words & 10.26 ($\pm$5.43) & 9.97 ($\pm$4.61) & 10.20 ($\pm$5.26) \\
         \bottomrule
    \end{tabular}
    \end{subtable}
\end{table}

\textbf{G{\textpipe}ui} (ISO 639-3: gwj) is a Kalahari Khoe language spoken in Botswana.
Until the late 1990s, G{\textpipe}ui speakers lived primarily within what is now the Central Kalahari Game Reserve (CKGR).
A large-scale survey conducted between 1996 and 1998 identified 769 speakers \cite{nakagawa-2006a-gui-dialects}.
No comprehensive census has since been carried out, and current estimates suggest fewer than 1,000 fluent speakers.
G{\textpipe}ui forms a close dialect cluster with G{\textdoublepipe}ana and is internally divided into three principal dialects (Xade, Tomelo, and Kute), distinguished primarily by consonantal features.
The phonological system of G{\textpipe}ui
exhibits the largest phoneme inventory reported within the Khoe--Kwadi family \cite{nakagawa-2006b-aspects-phonetic}.
The consonant system comprises approximately 90 phonemes, including 52 click consonants and 38 non-click consonants.
Four click types are distinguished: dental [{\textpipe}], alveolar [\textipa{!}], palatal [{\textdoublebarpipe}], and lateral [{\textdoublepipe}].
These combine with 13 click series (also referred to as accompaniment or efflux categories).
The series distinction is independent of click type and is defined by laryngeal articulations (voicing, aspiration, ejection/glottalization), oro-nasal processes, and posterior release modifications at the uvular place (see Table~\ref{tab:click-consonants}).
Linguists and the speaker community have been working on developing an orthography for G{\textpipe}ui \cite{kato-2025-gui-gana-dictionary}, which serves as the basis for the orthography used in the G|ui dataset for this paper.

The G{\textpipe}ui dataset consists of 50 audio recordings collected outdoors in New Xade (Ghanzi District), Botswana.
The recordings are narratives on G{\textpipe}ui folktales or personal experiences.
The dataset is not currently publicly available due to containing personally identifiable information and an incomplete agreement with the speech contributors on public release.
The dataset is split into the train set (90\%) and the test set (10\%), without a validation set due to its limited data size.

\begin{table}[t]
    \centering
    \caption{G{\textpipe}ui click consonants.
    For each cell, the symbols to the left show a phonetic value, and the bracketed symbols are orthographic notations used in the dataset.}
    \newcommand{\colw}[0]{4.3em}
    \setlength{\tabcolsep}{3.4pt}
    \label{tab:click-consonants}
    \begin{tabular}{@{}lp{\colw}p{\colw}p{\colw}p{\colw}@{}} \toprule
        Series & Dental & Alveolar & Palatal & Lateral \\ \midrule
        1. Plain & \dc ~\orth{\dc} & \ac ~\orth{\ac} & \pc ~\orth{\pc} & \lc ~\orth{\lc} \\
        2. Voiced & \ipag\dc ~\orth{\dc g} & \ipag\ac ~\orth{\ac g} & \ipag\pc ~\orth{\pc g} & \ipag\lc ~\orth{\lc g} \\
        3. Aspirated & \dc\h ~\orth{\dc h} & \ac\h ~\orth{\ac h} & \pc\h ~\orth{\pc h} & \lc\h ~\orth{\lc h} \\
        4. Ejective & \dc' ~\orth{\dc k'} & \ac' ~\orth{\ac k'} & \pc' ~\orth{\pc k'} & \lc' ~\orth{\lc k'} \\
        5. Nasal & \Ng\dc ~\orth{\dc n} & \Ng\ac ~\orth{\ac n} & \Ng\pc ~\orth{\pc n} & \Ng\lc ~\orth{\lc n} \\
        6. Plain+\X & \dc\X ~\orth{\dc x} & \ac\X ~\orth{\ac x} & \pc\X ~\orth{\pc x} & \lc\X ~\orth{\lc x} \\
        7. Plain+q\X' & \dc q\X' ~\orth{\dc x'} & \ac q\X' ~\orth{\ac x'} & \pc q\X' ~\orth{\pc x'} & \lc q\X' ~\orth{\lc x'} \\
        8. Plain+q & \dc q ~\orth{\dc q} & \ac q ~\orth{\ac q} & \pc q ~\orth{\pc q} & \lc q ~\orth{\lc q} \\
        9. Plain+\qg & \dc\qg ~\orth{\dc qg} & \ac\qg ~\orth{\ac qg} & \pc\qg ~\orth{\pc qg} & \lc\qg ~\orth{\lc qg} \\
        10. Plain+q\h & \dc q\h ~\orth{\dc qh} & \ac q\h ~\orth{\ac qh} & \pc q\h ~\orth{\pc qh} & \lc q\h ~\orth{\lc qh} \\
        11. Plain+q' & \dc q' ~\orth{\dc q'} & \ac q' ~\orth{\ac q'} & \pc q' ~\orth{\pc q'} & \lc q' ~\orth{\lc q'} \\
        12. Plain+\textipa{P} & \dc\textipa{P} ~\orth{\dc'} & \textipa{!P} ~\orth{\ac'} & \pc\textipa{P} ~\orth{\pc'} & \lc\textipa{P} ~\orth{\lc'} \\
        13. Plain+h & \dc h ~\orth{\dc nh} & \ac h ~\orth{\ac nh} & \pc h ~\orth{\pc nh} & \lc h ~\orth{\lc nh} \\
        \bottomrule
    \end{tabular}
\end{table}

\textbf{West \textipa{!}Xoon} (ISO 639-3: nmn) is a Tuu language spoken in Botswana and Namibia.
West \textipa{!}Xoon is the most widely used dialect of Taa that has around 3,000 speakers and is spoken in the region near Corridor 13 and Aminuis of Namibia \cite{naumann-2016-phoneme-inventory-taa}.
In addition to the four click types found in G{\textpipe}ui, a bilabial click [{\textbullseye}] is distinguished in West \textipa{!}Xoon.
Together with the accompaniments, the phonological inventory of West \textipa{!}Xoon has 43 click consonants, making it the most consonant-rich language in the Tuu language family.
The West \textipa{!}Xoon orthography is given in detail in Naumann (2016) \cite{naumann-2016-phoneme-inventory-taa}.

For West \textipa{!}Xoon, 150 minutes of elicited speech collected by scholars from the DoBeS program\footnote{https://dobes.mpi.nl} is used in the experiments.
The collection has been curated in a semi-automatic way, where speech--transcription matching has been assessed using the methods by Le Ferrand et al. \cite{le2025doesn}; then, mismatches are discarded, and shallow misalignment is manually corrected.
The dataset comprises approximately 1.75 hours of recording, which are then split into the train set (80\%) and the test set (20\%)

Both G{\textpipe}ui and West \textipa{!}Xoon are tonal languages.
G{\textpipe}ui has three level tones (High, Mid, Low) assigned to each mora, and West \textipa{!}Xoon distinguishes two level tones (High and Low).
In bimoraic roots, these level tones are combined to form a tonal melodies.
However, since the datasets contained both tone-aware transcription with diacritics and transcription without tones, the tone symbols (the grave accent U+0300, the acute accent U+0301, the macron U+0304) are removed.
Then, the text is lowercased and non-phonemic symbols such as parentheses are removed.

\section{Experiments}
Using the constructed datasets, we fine-tune the pretrained self-supervised models (Wav2Vec 2.0 series and HuBERT) to G{\textpipe}ui and West \textipa{!}Xoon and evaluate whether the models can adaptively learn to recognize the click consonants.
ASR models with autoregressive decoding are not compared here because their click recognition can be aided by the contextual information.
Table~\ref{tab:models} shows a list of the models compared in the experiments, as well as the number of languages included in their pretraining stage and the pretraining languages that have click consonants.
Note that, while Zulu and Xhosa are included in the pretraining of multilingual Wav2Vec 2.0-based models, certain click consonants such as palatal click [{\textdoublebarpipe}] (found in both G{\textpipe}ui and West \textipa{!}Xoon) and bilabial click [{\textbullseye}] (found in West \textipa{!}Xoon) never appear in these languages.

\begin{table}[t]
    \centering
    \caption{List of the models compared in the experiments.
    The column ``Size'' refers to each model's parameter size.
    ``\#lang'' shows the number of languages used in the pretraining.
    ``Click langs'' lists the languages with click consonants included in the pretraining data.
    The ISO 639-3 language codes \texttt{zul}, \texttt{xho}, \texttt{nde} refer to Zulu, Xhosa, and Northern Ndebele, respectively.}
    \label{tab:models}
    \setlength{\tabcolsep}{4pt}
    \begin{tabular}{@{}llll@{}} \toprule
        Model & Size & \#lang & Click langs \\ \midrule
        wav2vec2-large-xlsr-53 & 300M & 53 & \texttt{zul} \\
        wav2vec2-xls-r-300m & 300M & 128 & \texttt{zul} \\
        wav2vec2-xls-r-1b & 1B & 128 & \texttt{zul} \\
        mms-1b & 1B & $>$1,400 & \texttt{zul}, \texttt{xho}, \texttt{nde}, etc. \\
        mms-1b-all & 1B & $>$1,400 & \texttt{zul}, \texttt{xho}, \texttt{nde}, etc. \\
        hubert-large-ll60k & 300M & 1 & NA \\
        hubert-xlarge-ll60k & 1B & 1 & NA \\
        \bottomrule
    \end{tabular}
\end{table}

For mms-1b-all, an initialized adapter (language-specific linear projection layers for each attention block and a vocabulary output layer) is appended; for other models, an initialized vocabulary output layer is appended.
All the model training employs the same hyperparameters.
For the model configuration, attention dropout, hidden dropout, feature projection dropout, and layerdrop are set to 0.0, mask time probability to 0.05, and the CTC loss reduction method takes the mean over a batch.
Each training is run for 10 epochs with a learning rate of 0.0003 and a batch size of 8, optimized by the AdamW \cite{loshchilov2019decoupledweightdecayregularization}.
The first 100 steps are reserved as warm-up steps.
The models are trained on 24GB A10 GPUs.
Fine-tuning a 300M parameter model takes approximately 70 minutes.
We use Character Error Rate (CER) as the validation metric during training.
In addition, we report Phoneme Error Rate (PER) that treats multigraphs (\textit{e.g.}, \orth{kx'}) and complex clicks with an accompaniment (\textit{e.g.}, \orth{\ac qg}) as single symbols.

We measure the inference performance with four CTC decoding methods: greedy decoding, beam search decoding, beam search decoding with a 3-gram language model and with a 5-gram language model.
The beam width is set to 50.
The language models are obtained from the same training corpus using \texttt{kenlm} \cite{heafield-2011-kenlm}.
For all decoding with the language models, the hyperparameter $\alpha$ (language model weight) is set to 0.2 and $\beta$ (length penalty) to 0.0.
The integration of CTC with a language model is implemented using \texttt{pyctcdecode}\footnote{\url{https://github.com/kensho-technologies/pyctcdecode}}.

\section{Results}
\begin{figure}[ht]
    \centering
    \caption{PER across models and decoding methods. }
    \label{fig:PER_combined}
    \begin{subfigure}{\linewidth}
        \centering
        \caption{West \textipa{!}Xoon}
        \label{fig:PER_Xoon}
        \includegraphics[width=\linewidth]{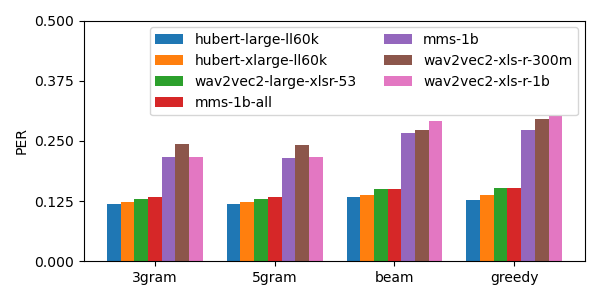}
    \end{subfigure}
    
    \vspace{0.2cm}
    
    \begin{subfigure}{\linewidth}
        \centering
        \caption{G{\textpipe}ui}
        \label{fig:PER_Gui}
        \includegraphics[width=\linewidth]{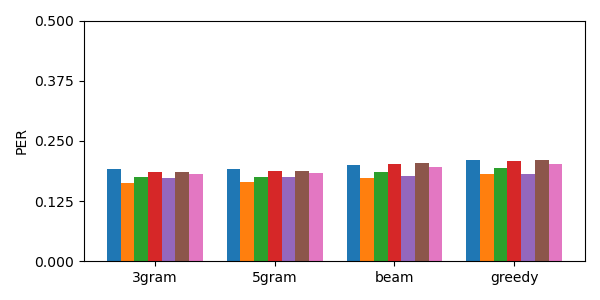}
    \end{subfigure}
    \vspace{-2em}
\end{figure}

\begin{figure}[ht]
    \centering
    \caption{Error rates of (non-)click consonants and vowels.
    }
    \label{fig:ClickER_combined}
    
    \begin{subfigure}{\linewidth}
        \centering
        \caption{West \textipa{!}Xoon}
        \label{fig:ClickER_Xoon}
        \includegraphics[width=\linewidth]{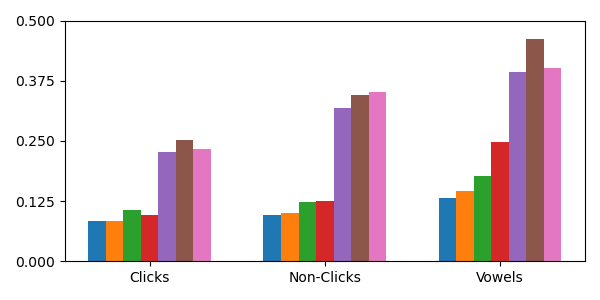}
    \end{subfigure}
    
    \vspace{0.2cm}
    
    \begin{subfigure}{\linewidth}
        \centering
        \caption{G{\textpipe}ui}
        \label{fig:ClickER_Gui}
        \includegraphics[width=\linewidth]{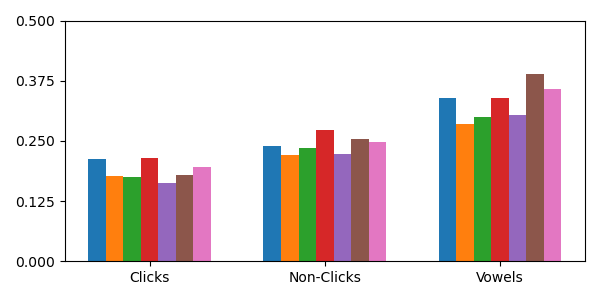}
    \end{subfigure}
    \vspace{-2em}
\end{figure}

This section demonstrates the comparative performance results across the various pretrained speech models on the ASR task of G{\textpipe}ui and West \textipa{!}Xoon.
The reported results are based on the checkpoint achieving the lowest CER on the evaluation set during training.
Figure~\ref{fig:PER_combined} shows the compared PERs of the models with varying decoding methods in the two languages.
Figure~\ref{fig:ClickER_combined} provides error rates with respect to different phoneme types (clicks, non-clicks, and vowels) across model.
The error rates, including PER, were computed by obtaining the optimal alignment between the reference and predicted transcriptions using the Needleman--Wunsch algorithm \cite{needleman1970general}.

\vspace{.4em}
\noindent

\noindent
\textbf{Model size does not matter.}
In both G{\textpipe}ui and West \textipa{!}Xoon, the larger parameter sizes did not necessarily provide better transcription performance.
wav2vec2-xls-r-300m often outperformed wav2vec2-xls-r-1b in both languages, and hubert-large-ll60k (300M) consistently beat hubert-xlarge-ll60k (1B) in West \textipa{!}Xoon.

\newcommand{\paragraphsep}[0]{.4em}
\vspace{\paragraphsep}
\noindent
\textbf{Base model parameters need to be updated.}
For mms-1b-all, we also fine-tuned it with its base model frozen, where only the adapter parameters are updated during training.
Our experiments revealed that the model did not perform well in G{\textpipe}ui when the base model parameters are frozen, with the PER nearly twice as high as the full-parameter fine-tuning setting.

\vspace{\paragraphsep}
\noindent
\textbf{Massively multilingual models can struggle.}
In addition, the experiments showed that pretraining on more languages did not guarantee better performance.
As shown in Figure~\ref{fig:PER_combined}, the models that performed best on the two languages were HuBERT models, which had been pretrained on monolingual English Libri-Light \cite{Kahn_2020}.
Among the Wav2Vec 2.0-based models, wav2vec2-large-xlsr-53, wav2vec2-xls-r-300m, and mms-1b all exhibited similar error rates in G{\textpipe}ui, as Figure~\ref{fig:PER_Gui} shows.
In West \textipa{!}Xoon, too, the monolingually pretrained HuBERT models performed the best overall, and wav2vec2-large-xlsr-53 yielded the best performance among the Wav2Vec 2.0-based models.

\begin{figure}[th]
    \centering
    \caption{PER by manner of articulation in G{\textpipe}ui.}
    \label{fig:per-by-manner}
    \includegraphics[width=1\linewidth]{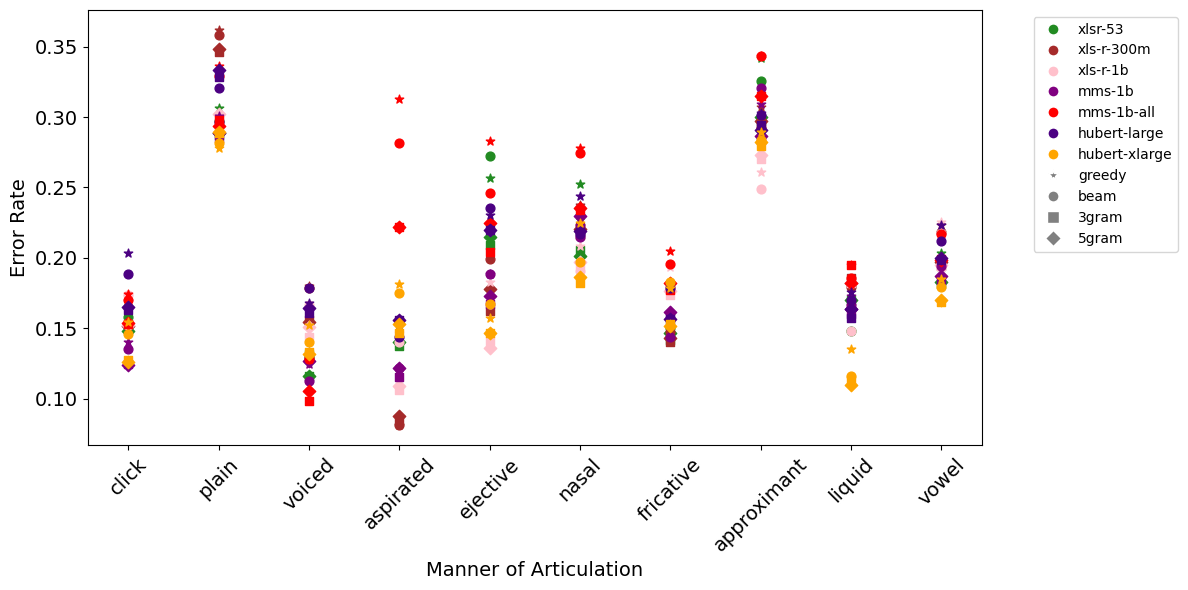}
    \vspace{-2em}
\end{figure}

\vspace{\paragraphsep}
\noindent
\textbf{Clicks are recognized more accurately than non-click phonemes.}
Despite the complex phonological system of the clicks, Figure~\ref{fig:ClickER_combined} shows that clicks were more accurately recognized than the other consonants and vowels, and Figure~\ref{fig:per-by-manner} shows in detail that clicks are recognized relatively better and more stably than the other manners of articulation.
A Wilcoxon signed-rank test comparing error rates of click consonants and non-click phonemes under greedy decoding shows that click consonants are recognized with significantly lower error rates ($W = 0$, $p = 0.016$).
This observation is in line with the report that the performance of self-supervised pretrained speech models is robust to phonological complexity \cite{taguchi-chiang-2024-language}.
This also suggests that the self-supervised pretraining methods of Wav2Vec 2.0 and HuBERT enable recognition of unseen or rarely seen click consonants through full-parameter fine-tuning.
This result is consistent with phonetic research indicating that click consonants are perceptually more salient and identifiable than non-click consonants \cite{ladefoged-traill-1994-clicks-accompaniments}.

Figure~\ref{fig:ClickER_combined} also shows that vowels exhibit relatively higher error rates in both languages.
A preliminary inspection of the vowel errors suggests that most confusions occur among oral, nasalized, and long vowel realizations.
For example, among all errors involving the phoneme /a/, 21\% are recognized as /aa/ and 8\% as /an/.
Unlike consonantal contrasts, vowel quality distinctions such as length, nasality, and phonation often form more gradient acoustic continua, which may make them more difficult for ASR models to distinguish reliably.

\section{Conclusion}
This study investigated whether self-supervised pretrained speech models are able to recognize click consonants using newly constructed ASR datasets for two Khoisan languages, G{\textpipe}ui and West \textipa{!}Xoon.
Although click consonants are extremely underrepresented in the pretraining data of existing self-supervised speech models, these models did not exhibit particularly low performance on these click-rich languages.
In-depth analyses further revealed that
(1) larger models (1B params) did not necessarily produce more accurate transcriptions than smaller models (300M params), and
(2) models pretrained on a larger number of languages did not consistently yield better results than those pretrained on fewer languages, and were sometimes even outperformed by the monolingually pretrained HuBERT models.
Importantly, these models consistently yielded lower phoneme error rates for click consonants than for non-click phonemes.
This finding suggests that the self-supervised training paradigm of these speech models exhibits strong adaptability to speech sounds unseen in the pretraining data.

\section{Acknowledgments}
The material was based on work supported in part by the US National Science Foundation under Grant Number BCS-2109709 and IIS-2137396 and by the JSPS KAKENHI Grant Number 22H04929, 22K18249, 23K25318, and 22K00536.
We thank Florian Lionnet Alena Witzlack-Makarevich for the West \textipa{!}Xoon data.
We are also grateful to the reviewers and the meta-reviewer of Interspeech 2026 for providing constructive feedback.

\section{Generative AI Use Disclosure}
The first author of the paper, whose first language is not English, used generative AI tools to check grammar and improve the flow of the writing.
Code autocompletion tools were also used for building the experimental code.
All AI-generated content was carefully reviewed by the author and by native English speakers.
The authors take full responsibility for the content of this paper.

\bibliographystyle{IEEEtran}
\bibliography{mybib}

\end{document}